\documentclass[10pt, a4paper]{article}

\usepackage[]{lrec-coling2024} 

\usepackage{multirow}

\usepackage {xcolor}
\usepackage{amsmath}
\usepackage{amssymb}
\usepackage{graphicx}
\usepackage{url}
\usepackage{subfig}
\usepackage{authblk}

\title{ASEM: Enhancing Empathy in Chatbot through Attention-based Sentiment and Emotion Modeling}

\name{Omama Hamad\textsuperscript{1}, Ali Hamdi\textsuperscript{2}, Khaled Shaban\textsuperscript{1}} 

\address{\textsuperscript{1}Qatar University, Department of Computer Science and Engineering, Qatar\\
         \textsuperscript{2}MSA University, Department of Computer Science, Egypt\\
         \{omama.hamad, khaled.shaban\}@qu.edu.qa, ahamdi@msa.edu.eg\\}

\abstract{
Effective feature representations play a critical role in enhancing the performance of text generation models that rely on deep neural networks. However, current approaches suffer from several drawbacks, such as the inability to capture the deep semantics of language and sensitivity to minor input variations, resulting in significant changes in the generated text. In this paper, we present a novel solution to these challenges by employing a mixture of experts, multiple encoders, to offer distinct perspectives on the emotional state of the user’s utterance while simultaneously enhancing performance. We propose an end-to-end model architecture called ASEM that performs emotion analysis on top of sentiment analysis for open-domain chatbots, enabling the generation of empathetic responses that are fluent and relevant. In contrast to traditional attention mechanisms, the proposed model employs a specialized attention strategy that uniquely zeroes in on sentiment and emotion nuances within the user's utterance. This ensures the generation of context-rich representations tailored to the underlying emotional tone and sentiment intricacies of the text. Our approach outperforms existing methods for generating empathetic embeddings, providing empathetic and diverse responses. The performance of our proposed model significantly exceeds that of existing models, enhancing emotion detection accuracy by 6.2\% and lexical diversity by 1.4\%. ASEM code is released at \href{https://github.com/MIRAH-Official/Empathetic-Chatbot-ASEM.git}{https://github.com/MIRAH-Official/Empathetic-Chatbot-ASEM.git.}
 \\ \newline  }

\begin{document}

\maketitleabstract

\section{Introduction}
The ability of deep learning models to generate natural language is highly dependent on the quality of the feature representations, a.k.a., embeddings, that capture the underlying patterns in text. Thus, representation learning is crucial for various natural language processing tasks, including text generation, which mimics a human skill. However, this task can be challenging due to colloquial dialects and the diverse ways emotions are expressed, blurring the boundaries between distinct emotional states. Additionally, certain emotions, such as surprise and fear, exhibit close relationships as evidenced by psychological studies, which can lead to confusion in the model's understanding \citet{noordewier2013valence}. Consequently, developing high-quality text generation models requires complex architecture, extensive data resources, and an understanding of linguistic nuances. Recent advancements have been made in utilizing different techniques to enhance and learn representations \citet{wu2021effective, el-boukkouri-etal-2022-specializing}. 

\begin{figure}
  \centering
  \includegraphics[width=1\linewidth]{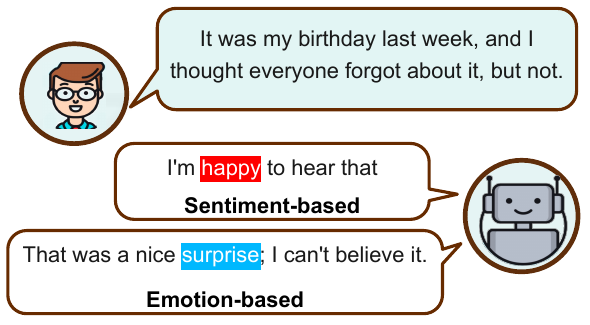}
  \caption{Emotional vs Sentimental Response.}
  \label{Emotion vs sentiment}
\end{figure} 
In this work, we propose an effective attention-based model for empathetic text generation in open-domain chatbots, amining to engage and interact with users in a natural, human-like manner. Furthermore, the model is trained with empathetic conversations to create a more immersive user experience and align with their emotional states. Previous studies have employed sentiment or emotion modelling to address empathetic conversational abilities \citet{zaranis2021empbot, majumder-etal-2020-mime}. Sentiment refers to the overall tone or mood of a user's utterance, while emotion represents specific feelings or mental states, encompassing over 30 emotions. Emotions are often described along dimensions like valence (positive or negative), arousal (high or low), and dominance (high or low) \citet{plutchik1980general}. Although sentiment analysis offers insights into the overall tone, it may not capture the full spectrum of emotions a user experiences. On the other hand, emotion analysis helps identify specific emotions, such as joy, fear, or surprise. 

\begin{figure}
  \centering
  \includegraphics[width=1\linewidth]{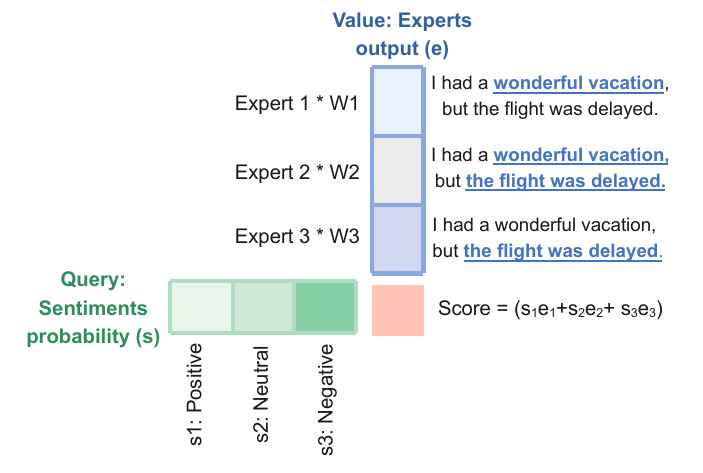}
  \caption{Attention Weights.}
  \label{Attention Weights}
\end{figure}

Consequently, relying solely on sentiment analysis may cause confusion in the generated embeddings when texts with the same sentiment belong to different emotions. For instance, statements like "That was a nice surprise" and "I'm happy to hear that" may express similar sentiment (positive) but convey different emotions (surprise and happiness) as depicted in Figure \ref{Emotion vs sentiment}. This highlights the importance of emotion analysis in understanding the user's feelings and generating appropriate responses. Hence, we incorporate emotion analysis on top of sentiment analysis using a mixture of experts to enable the empathetic chatbot to comprehend the user's emotional state more comprehensively. This approach facilitates the generation of empathetic responses tailored to the user's specific emotional needs. For example, if a user expresses sadness in their message, sentiment analysis may detect a negative sentiment, but emotion analysis can identify the specific emotion of sadness. Consequently, the chatbot can then generate a response that acknowledges the user's sadness and provides words of comfort or support. Introducing a pioneering attention mechanism, the proposed model uniquely discerns and prioritizes sentiment- and emotion-laden aspects of a user's utterance, ensuring the generation of representations that deeply capture the intricate contextual nuances within the text. The proposed approach enables the chatbot to provide a score indicating how well the generated embedding matches the user's sentiment. Additionally, the magnitude of each sentiment serves as an indicator of how strongly each expert should contribute to the total score. For example, as depicted in Figure \ref{Attention Weights}, a statement may exhibit a combination of positive, negative, and neutral sentiments.

Our contributions can be summarized as follows:
\begin{itemize}
    \item Sentiment and Emotion Experts: We propose a comprehensive model for open-domain chatbots that can recognize the sentiment and emotion in user statements, enabling the generation of empathetic responses tailored to the user context.

    \item The use of a specialized attention strategy: We compute the score of each expert based on the proposed attention mechanism, improving the model's ability to generate empathetic responses. This improvement was demonstrated through both automatic and human evaluations.
    \item Standardization of emotion categories: We establish a standardized set of emotion categories across two datasets, enhancing consistency and comparability in emotion analysis.
    

\end{itemize}

Overall, our proposed ASEM model enhances the chatbot's empathetic capabilities, providing more accurate sentiment and emotion recognition, and generating contextually relevant responses that resonate with users' emotional states.

\section{Related Work} \label{rw}
Open-domain chatbots are AI-based systems designed to engage and entertain humans in conversational interactions. These chatbots require various skills to effectively emulate human conversations, and empathy is considered one of the most important skills. Empathy involves understanding and being attuned to the emotional states of users. Psychological studies have identified different methods through which humans express empathy \citet{baron2006empathy, low2012being, song2015proponent}. These methods include 1) expressing care and concern, 2) acknowledging the user's feelings to avoid closed-off responses such as ‘ok’ or ‘I see’, 3) asking questions that encourage sharing of feelings and 4) demonstrating emotion-supporting behaviour. Each of these cases is illustrated in Table \ref{Empathetic cases}. However, existing research that relies on human evaluations of empathetic scenarios often does not specify the level of empathy involved, ranging from weak to robust. Therefore, we propose a framework based on these four cases for human evaluation of empathetic responses generated by the chatbot.

Recent research has explored various perspectives in this domain. For example, some studies focused on understanding sentiments in user inputs to generate responses with a sentimental tone \citet{zaranis2021empbot}. \citet{firdaus-etal-2021-seprg} proposed a transformer-based model for generating emotional responses aligned with the user's persona and conditioned on a specific sentiment. However, relying solely on sentiment may not always yield the desired emotion, as emotional analysis is more comprehensive \citet{kumar2019emotion}. Multitask architectures have been employed to jointly learn and classify sentiments and emotions, utilizing attention mechanisms to incorporate information from various sources \citet{akhtar-etal-2019-multi}. Another recent study introduced a multi-party empathetic dialogue generation task, employing a dynamic graph network to incorporate temporal information and blend dynamic emotions and static sensibilities from different parties \citet{zhu2022multi}. Temporal and spatial information can be crucial for empathetic text generation, enabling the understanding of local entity relationships and interactions. Additionally, there have been approaches based on fine-grained or coarse-grained emotion analysis \citet{lin-etal-2019-moel, naous-etal-2020-empathy}. For example, open-domain chatbots like XiaoIce have demonstrated the ability to understand users' emotional needs, but their generated responses may not always be accurate and rely on external knowledge \citet{zhou2020design}. Other end-to-end approaches have encountered challeges in generating irrelevant or low-diversity responses \citet{zhang-etal-2020-dialogpt}.

In summary, while existing research on open-domain chatbots has made significant strides in understanding and generating empathetic responses, a clear gap remains in the nuanced application and evaluation of empathy in these interactions. Current studies often do not define the varying levels of empathetic engagement, ranging from basic acknowledgment to deep emotional resonance, which are crucial for truly human-like conversational experiences. Moreover, the reliance on sentiment analysis alone has proven insufficient for capturing the full spectrum of human emotions and their context within conversations. Our proposed work aims to bridge this gap by introducing a more structured and comprehensive approach to evaluating empathetic responses in chatbots. By focusing on a multi-dimensional evaluation that encompasses both the depth of empathy and the integration of emotion analysis on top of sentiment analysis to increase and accurately acknowledge and understand empathy, we intend to enhance the empathetic capabilities of chatbots, making them more adept at navigating complex emotional landscapes in human interactions.

\section{Methodology}

\subsection{Problem Formulation}
In multi-turn conversational settings, the problem is formalized as the exchange between a user’s statement $U$ and a chatbot’s response $R$, represented by the overall user context as $C = \{U_{1}, R_{1}, U_{2}, R_{2},...\}$. The emotion of the user is expressed in each user statement, such as joy or surprise, and belongs to a sentiment class, which can be positive, negative or neutral, denoted as $E = \{e_{j}, e_{l}, ..,e_{n}\}$ and $S = \{s_{pos}, s_{neg}, s_{neu}\}$, respectively. The user's history includes all previous exchanges and context, but excludes the most recent statement by the user. The objective of the model is to identify the sentiment and emotion of the user’s statement based on the current statement and the conversation history $C$, and generate a relevant, coherent and empathetic response.

\begin{table}
\centering
\small
\begin{tabular}{ll}
\hline
\textbf{Context} &
  \begin{tabular}[c]{@{}c@{}}
  I'm having to cancel my Labor Day plans\\ since I have to work. I'm disappointed.   \\
  \end{tabular}
\\
  \hline
\multicolumn{1}{l}{} & \multicolumn{1}{c}{\textbf{Responses}}                              \\ \hline
\multirow{1}{*}{\textbf{Case 1} }       &  How do you feel now? \\

\multirow{2}{*}{\textbf{Case 2}}   &   I'm sorry to hear that. I realise that what\\ &you are going through is very difficult. \\

\multirow{2}{*}{\textbf{Case 3}}       &  
Oh no, I'm so sorry to hear that. Is there \\ &anyone who can help you?
\\

\multirow{2}{*}{\textbf{Case 4}}       & 
I'm sorry to hear that. Is there anything I \\ &can do for you? \\

\hline
\end{tabular}%
\caption{Empathetic Cases.}
\label{Empathetic cases}
\end{table}

\begin{figure*}
  \centering
  \includegraphics[width=0.9\linewidth]{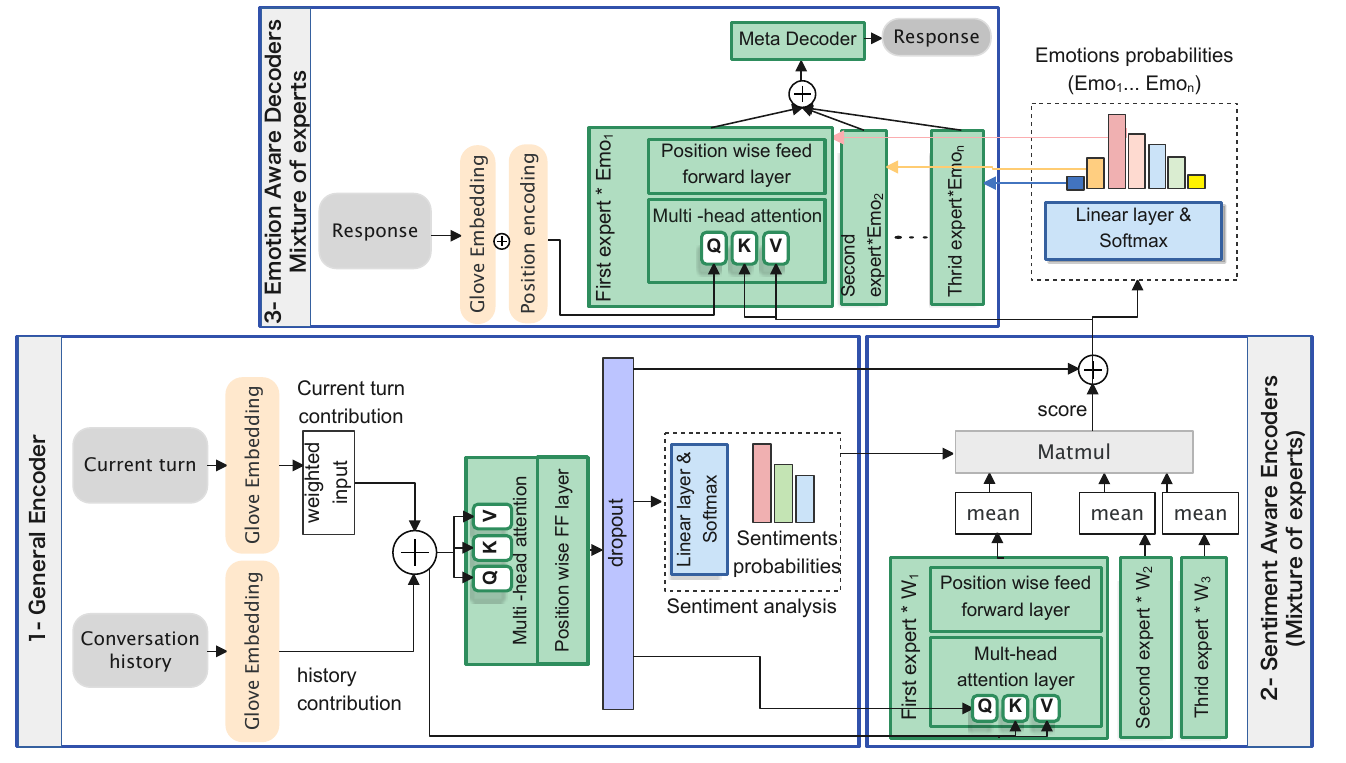}
  \caption{ASEM model for empathetic chatbot.}
  \label{proposed empathetic chatbot}
\end{figure*}

\subsection{Proposed Model} \label{proposedMod}
We propose a novel approach that sequentially tracks emotions and sentiments to create a more effective and nuanced chatbot. Our proposed method, illustrated in Figure \ref{proposed empathetic chatbot}, consists of two stages: learning empathetic feature representations and generating an empathetic response. The first stage includes a sentiment-tracker and an emotion tracker, while the second stage adopts the concept of a mixture of empathetic listeners from \citet{lin-etal-2019-moel} for generating an empathetic response. A key novelty in our approach is the mechanism to combine the outputs of these experts. Instead of directly pooling or averaging the experts' outputs, our model computes the mean of all expert outputs and multiplies them with the sentiment probability distributions. This ensures that each sentiment-specific feature is weighted appropriately based on its relevance, producing a sentiment-specific context-rich feature representation. Furthermore, by adding the weighted feature representation with the general encoder's output, the model not only retains the original semantic information but also infuses it with sentiment-aware nuances. This enriched representation is then used to predict emotion classes, thereby grounding the model's attention mechanism in both semantic and emotional dimensions of the text.


 


\subsubsection{Stage 1: Learning Empathetic Features Representation}

\textbf{General Encoder: }
The general encoder is responsible for generating feature representations of sequences $G_{H}$ with corresponding probability distributions for different sentiment classes $S_{att}$. These probability distributions are used as attention weights for the next group of experts. The general encoder takes two inputs: the current user turn $U$ and the previous conversation history $C$, both represented by GloVe embeddings. The encoder consists of multi-head attention and position-wise feed-forward layers. 

To control the importance of the current turn compared to the conversation history $U_{w}$, a weighting factor is applied to the embeddings of the current turn. This weighting factor allows the model to learn to focus more on recent information while still considering the context provided by the overall conversation history. The weighted representation of the current turn is then combined with the conversation history and fed into the encoder:

\begin{equation}
U_{w} = Emb(U) \times W
\end{equation}

\begin{equation}
G_{h} = Enc([U_{w}, Emb(C)])
\end{equation}
Here, $G_h \in \mathbb{R}^{L \times d}$ represents the last hidden state of the general encoder, where $L$ is the length of the conversation and $d$ is the hidden dimension of the encoder.

To obtain the probability distribution $S_{att}$, a softmax function is applied to $G_H$:
\begin{equation}
S_{att} = Softmax(G_H)
\end{equation}

The performance of the general encoder is optimized by calculating the loss using the cross-entropy function, which measures the difference between the predicted sentiment label $p_{s}$ and the actual label $y_{s}$. 

\begin{equation}
L_{1} = -y_{s} \log \left(S_{att}\right)
\end{equation}

 

\textbf{Sentiment Aware Encoders (Mixture of Experts): }
Sentiment Aware Encoders (SAEs) are designed to generate sentiment-specific feature representations that capture the contextual nuances of the input text. Unlike conventional encoders that produce fixed representations, SAEs employs a set of encoders, each initialized with different weights and dedicated to modeling a specific sentiment. The last hidden state of the general encoder, denoted as $G_{h}$, serves as query input to each encoder in the attention layer, while the concatenated input $C$ is the key and value. The context of each input is captured by taking the mean of the output tensors of each expert, allowing us to combine the outputs and generate a more informative representation $S_{h}$ of the input text. By performing a batch matrix multiplication between the sentiment probability distributions $S_{att}$ and the output of the expert mean tensor, we obtain a sentiment-specific context-rich feature representation. This computation can be performed efficiently by computing the weighted sum for all examples in the batch in a single operation.


\begin{equation}
S_{h} =\overline{E}_{0}((G_{h}, Emb(C)) * W_{i}),..., \overline{E}_{i}
\end{equation}
where $\overline{E}_{i}$ represents the output of the $i-th$ expert encoder and $W_{i}$ represents the weight matrices for the i-th encoder.


To weight the output features $S_{h}$ based on their relevance, the attention score $W_{att}$ is calculated. This score is computed as the weighted sum of the outputs of the three experts $k$:

\begin{equation}
    W_{att_{i,j,l}} = \sum_{k} S_{att_{i,j,k}} \times S_{h_{i,k,l}}
\end{equation}

Here, the attention-weighted output \( W_{att_{i,j,l}} \) for each data instance \( i \) in the batch, feature \( j \), and hidden dimension \( l \) is obtained by summing over the product of attention scores \( S_{att_{i,j,k}} \) and the corresponding expert outputs \( S_{h_{i,k,l}} \) for all experts \( k \). The attention scores serve as weights that determine the significance of each expert’s contribution, effectively allowing the model to focus on the most relevant features provided by each expert for constructing the final output representation.


Next, the weighted input features $W_{att}$ are combined with the general encoder output $G_{h}$:

\begin{equation}
S_{h} = W_{att} + G_{h}
\end{equation}

To perform emotion analysis, a probability distribution over seven emotion classes is generated, indicating the likelihood of the input context $C$ belonging to each emotion class $E_{att}$. This distribution quantifies the degree of association between the input context and each emotion class.

\begin{equation}
    E_{att} = Softmax(S_{H})
\end{equation}

To improve the performance of the sentiment-aware encoders, the cross-entropy loss is calculated between the predicted emotion label $E_{att}$ and the actual emotion label $y_{e}$.

\begin{equation}
L_{2} = -y_{e} \log \left(E_{att}\right)
\end{equation}

\subsubsection{Stage 2: Generating Empathetic Response}

In this phase, we adopt the techniques of a mixture of empathetic listeners \citet{lin-etal-2019-moel} to generate an empathetic response. The objective is to train each decoder, represented as $D_{i}$, to respond to the input context with a specific emotion. These decoders consist of standard decoder layer blocks and take the response embedding $Emb^{R}$ as the query, and the output of the sentiment-aware encoders $S$ as key and value parameters. Each decoder is assigned a weight based on the probability distributions $E_{att}$. The outputs of all decoders are then combined and fed into the meta-decoder $D_{M}$, which aggregates multiple perspectives from emotion-based decoders to generate a response that appropriately expresses the desired emotion.

\begin{equation} 
    D_{i} = Dec(Emb^{R}, S_{h})
\end{equation}

\begin{equation}
    D_{M} = Dec(\sum_{i=1}^{n} D_{i} E_{att})
\end{equation}

The negative log likelihood loss is computed by considering the loss between the predicted words with the true words.

\begin{equation}
L3 = -\log \left(p(W | R) \right)
\end{equation}

Since all components of the proposed model are trained end-to-end, the gradient is computed based on the total of the three loss functions to minimise the sum of individual task losses.

\begin{equation}
L = L_{1} + L_{2} + L_{3}
\end{equation}


\section{Experimental Setup}

\subsection{Datasets}
The experimental setup involves training the proposed model on two datasets: the English Empathetic Dialogue (ED) dataset and the DailyDialog (DD) dataset \citet{rashkin-etal-2019-towards, li-etal-2017-dailydialog}. ED and DD consist of approximately 25,000 and 12,000 multi-turn conversations, respectively. The ED dataset provides annotations for 32 emotions, which we map to the eight basic emotions and two complex emotions defined by Plutchik's wheel of emotions \citet{plutchik1980general}. Appendix \ref{sec:appendixDataset} provides additional statistics on the datasets. 

DD dataset, on the other hand, is already labeled with the six main emotions defined by Plutchik's theory: anger, disgust, fear, happiness, sadness, surprise, and no emotion. By using Plutchik's Wheel of Emotions, we adopt a more manageable framework for modeling emotions compared to considering the 32 emotions individually. This allows for a standardized set of emotion categories across datasets and applications, facilitating the integration of both ED and DD datasets. Complex emotions, which are combinations of basic emotions, are also considered. For example, joy and trust combine to form love, while sadness and disgust combine to form remorse. 

By incorporating complex emotions, the model can capture a broader range of emotions, enhancing its sensitivity and the chatbot's ability to provide empathetic and supportive responses. The chatbot can learn to identify emotions that are actually a mix of two or more. In addition, emotions intensify as they move from the outside to the center of the wheel of emotions. For instance, anger ranges from annoyance to rage. As a result, we aggregate all the levels into its basic class. For example, emotions such as "grateful," "excited," and "content," are mapped into the class "joy." 

Using Plutchik's emotions provides several advantages. It assists in standardizing emotion categories across datasets. Incorporating complex emotions such as love and remorse into emotion classification may present challenges in prediction accuracy since these emotions may be wrongly labeled as other emotions such as joy or sadness. However, incorporating such emotions broadens the spectrum of emotions captured and improves the chatbot model's sensitivity. This enables chatbots to provide more empathetic and supportive responses, increasing user satisfaction and engagement. 

To handle the multi-dimensionality and contextual factors influencing emotions, emotion analysis is performed on top of sentiment analysis. In the ED dataset, emotion classes are classified as positive or negative. Since the DD dataset contains the "no emotion" class, the classes have been divided into positive, negative, and neutral sentiment classes\footnote{Preprocessed data is available at\url{}}. This approach helps disambiguate emotionally charged emotions, such as love, which may express both positive and negative emotions. By incorporating sentiment analysis into emotion classification, the overall accuracy of emotion analysis is improved.

\subsection{Implementation Details}
We utilized a virtual instance hosted on the Google Cloud Platform. This instance was equipped with 40 GB of RAM and a single NVIDIA A100 GPU unit. The number of trainable parameters in the model is 74 million. Our model's training took approximately 20 hours to complete. Our approach involved leveraging preexisting code as a basis and building upon it to develop our solution using the PyTorch framework \citet{lin-etal-2019-moel}. We used a batch size of 16 and implemented the early stopping strategy. In addition, batch matrix multiplication is employed to improve the model's efficiency. To initialize the embedding, we used transfer learning techniques by employing 300-dimensional English-pretrained Glove embeddings \citet{pennington2014glove}. The model was optimized with the AdamW optimizer and trained for about 20K steps. Each encoder and decoder consists of 12 layers, with 10 attention heads per layer. In addition, we assigned a weight of 2.5 to the current turn to prioritize its importance. The weight matrices are initialized using Xavier initialization. For more implementation details of our ASEM model, please refer to Appendix \ref{sec:appendixA}


\subsection{Baselines}
Three recent models are used as baselines for comparison with the ASEM model: \textbf{(1) CASE} \citet{zhou-etal-2023-case} model aligns cognition and affection in empathetic dialogue generation by leveraging a commonsense cognition graph and an emotional concept graph. It integrates three stages: graph encoding, coarse-to-fine alignment, and an empathy-aware decoding. \textbf{(2) Mixture of empathetic listeners (MoEL)} \citet{lin-etal-2019-moel} model is trained on the ED dataset, which includes 32 emotion classes, with the goal of generating empathetic responses \citet{rashkin2018know}. It uses a mixture of decoders, where the model attends to a specific decoder to generate an appropriate empathetic response. \textbf{(3) Multitask Transformer (MultiTRS)} \citet{rashkin2018know} model is based on a transformer architecture and incorporates an additional component for emotion classification. This enables the model to classify encoded information into specific emotions and generate a response based on the encoded information.



\subsection{Evaluation Metrics}

For the evaluation of chatbots, we followed the framework introduced in \citet{sedoc-etal-2019-chateval} for conducting automatic evaluation, and the framework introduced in \citet{sabour2022cem} for conducting human evaluation. We evaluate the performance of our model using the following metrics: \textbf{BLEU score} \citet{papineni2002bleu}: It measures the similarity between the generated responses and the reference responses using n-gram precision. \textbf {Lexical diversity (distinct-n)}: It calculates the average number of unique n-grams in the generated responses, which indicates the diversity of vocabulary used. \textbf{Average cosine similarity}: It measures the semantic similarity between the embeddings of the generated responses and the reference responses \citet{liu-etal-2016-evaluate}. \textbf{Perplexity}: It quantifies the probability of the model predicting the correct response \citet{zhang-etal-2018-personalizing}. 

In addition, we use the macro-F1 score metric to evaluate the emotion classification. Due to imbalanced datasets, we exclude the "no emotion" class when evaluating the performance of the proposed model fine-tuned on the DD dataset. This exclusion is necessary because the dataset contains a large number of dialogues without emotion compared to other classes. While BLEU score is commonly used in chatbot evaluation, it may not be reliable for open-domain response generation since the gold response is not necessarily the only correct response. Therefore, we report the BLEU score as a reference \citet{liu-etal-2016-evaluate}. 

In our experiments, we adopted a sequential analysis, first focusing on sentiment analysis before transitioning to emotion analysis. This strategic progression allowed us to harness sentiment data to dynamically refine attention parameters for detecting emotions, thus providing a nuanced insight into their interconnected nature. On the other hand, the joint learning framework implies a simultaneous learning process where both sentiment and emotion adjustments to attention parameters are conducted in an integrated manner, without the explicit, step-by-step dependency featured in the sequential approach. While joint learning presents benefits in concurrent sentiment-emotion recognition, our sequential methodology stands out, emphasizing sentiment-driven adjustments in attention for enhanced emotion discernment, ultimately yielding superior outcomes.

For human evaluation, we conducted an aspect-based pairwise preference test where we randomly sampled 135 dialogues from the ED test set and 100 from the DD test set. For a given context, we paired our model's response with a response from the baselines and presented them to three annotators. The annotators were asked to choose the better response based on the context and three aspects: coherence (Coh.) to assess which response was more coherent in content and relevant to the context, empathy (Emp.) to assess which response showed more understanding of the user's situation and presented a more appropriate emotion, and fluency (Flu.) to evaluate the response's flow and grammar. When both responses are equally good, then the annotator will select "it's a tie". In terms of calculating wins, losses, and ties, we considered the judgements of all three annotators for each of the samples. That is, for each sample, we have three separate evaluations. This method ensures a comprehensive evaluation by considering the independent judgement of each annotator on every sample, as our primary aim was not to achieve consensus but to understand the diverse perspectives of individual annotators.


\begin{table}[]
\centering
\resizebox{\columnwidth}{!}{%
\begin{tabular}{lcccccl}
\hline
\multicolumn{1}{l}{\textbf{Model} }
 &  
  \textbf{PPL} &
  \textbf{BLEU} &
  \textbf{D-1} &
  \textbf{D-2} &
  \textbf{Cos.} &
  \textbf{F1} \\
\hline
 \multicolumn{7}{c}{\textbf{ED dataset}}                                                     \\ \hline
  
MoEL                      & 54.9          & \textbf{2.8} & 38.6          & \textbf{6.7} & \textbf{24.7} & 54.4          \\
MultiTRS                  & 43.4 & 2.4          & 30.0          & 4.7          & 23.5          & 52.1          \\
 CASE& \textbf{36.7}& 2.0& 29.1& 2.0& 23.5& 52.3\\
 ASEM(j)                      & 42.7         & 1.63          & 38.7 & 2.6 & 22.4          & 46.4
 \\
ASEM(s)                       & 48.2          & 2.1          & \textbf{40.0} & \textbf{6.7} & 24.3          & \textbf{60.6} 
\\ \hline
\multicolumn{7}{c}{\textbf{DD dataset}}                                                                                 \\ \hline
MoEL &
 47.4 &
 1.6 &
27.1 &
\textbf{11.2} &
18.3 &
  34.2 \\
MultiTRS &
43.8 &
1.7 &
36.9 &
9.1 &
17.5 &
  16.3 \\
CASE & 57.8& 0.51& 42.1& 0.68& 13.4& 30.0  \\
ASEM(s)  &
\textbf{43.4} &
\textbf{2.2} &
\textbf{48.8} &
9.8 &
\textbf{18.4} &
  \textbf{39.2} 
 \\   \hline
\end{tabular}%
}
\caption{Automatic Evaluation results. Here PPL denotes perplexity, D-1 and D-2 denote Distinct-1 and -2, Cos. denotes the cosine similarity. 'ASEM (s)' refers to sequential training and 'ASEM (j)' to joint training.}
\label{tab:Results-final}
\end{table}

\section{Results and Analysis}\label{Results}
\subsection{Automatic Evaluation}
As shown in Table \ref{tab:Results-final}, the ASEM model outperforms baselines in three metrics using the ED and five metrics using the DD datasets. The results suggest that the ASEM model is able to produce a larger variety of unigrams compared to the baselines. However, when considering bigrams, our model generates comparable unique bigrams using the ED dataset, but it does not generate as diverse or unique combinations as the other baselines using the DD dataset, which could result in a higher repetition of word pairs. In addition, the improvement in PPL using the DD dataset indicates that the ASEM model is more confident in predicting a given sequence of words and is able to generate more coherent and fluent responses that are more closely aligned with human-like language usage. 
Regarding emotion classification using the macro F1 score, the ASEM model significantly outperforms the baselines by improving performance by $6.2\%$ and $5\%$ using the ED and DD datasets, respectively. This improvement suggests that incorporating emotion analysis on top of sentiment analysis unveils the confusion between various emotion classes that can be conveyed through both negative and positive sentiments. Additionally, the findings imply that the model learns an effective representation of the input sequence by employing multiple expert encoders and adding the attention score. By incorporating this score, the model enhances emotion prediction. ASEM improves performance in eight of ten emotion classes significantly. It also removes the confusion between more emotion classes of different sentiments, such as "surprise," which was never predicted as "remorse," and "disgust," which was never predicted as "joy," while the MoEL struggles with emotion classes of different sentiments. This indicates ASEM's capacity to minimize both false positives and false negatives and capture patterns and contextual cues that, in turn, help improve feature representations and generate more empathetic responses (Appendix \ref{sec:More Analysis} shows confusion matrices for each model).

\subsection{Human Evaluation}
As shown in Table \ref{tab:HumanEvalResults}, the human A/B test confirmed that our model's responses are preferred by human judges, indicating that the ASEM model outperforms the baselines in all three aspects. Fleiss' kappa measures inter-annotator agreement, with values of $\kappa$ > 0.4 and $\kappa$ > 0.2 indicating moderate and fair agreement, respectively.

\begin{table}
\centering
\resizebox{\columnwidth}{!}{%
\begin{tabular}{llllll}
\hline
\multicolumn{1}{c}{\textbf{Comparisons}} &

    \multicolumn{1}{c}{\textbf{Aspect}} &
  \multicolumn{1}{c}{\textbf{Win}} &
  \multicolumn{1}{c}{\textbf{Lose}} &
  \textbf{Tie }&
  \multicolumn{1}{c}{\textbf{$\kappa$}} \\ \hline
 &
  \multicolumn{5}{c}{\textbf{ED dataset}} \\ \hline

\multirow{3}{*}{ASEM vs. MoEL} &
  Coh. &   \textbf{45.4} &  21.2 & 31.1 & 0.41
 \\
& Emp. & \textbf{37.3}     & 22.5 & 40.0 &    0.44      \\
 & Flu. &    \textbf{11.4}       & 3.7 &  84.7 &   0.45        \\ \hline
 
\multirow{3}{*}{ASEM vs.MultiTRS} &
  Coh. & \textbf{51.1}           & 18.0 & 30.9 & 0.47
\\
& Emp. &    \textbf{48.1}       & 17.5 & 34.3 &   0.45      \\
& Flu. &     \textbf{12.1}      & 5.4 & \textbf{82.5} &    0.46       \\ \hline

\multirow{3}{*}{ASEM vs. CASE}    & Coh. &       \textbf{52.0}  & 15.0 &  41.0& 0.55 \\
                                     & Emp. &     \textbf{71.0}    & 39.0 & 25.0&   0.57     \\
                                     & Flu. &  \textbf{42.0}  &  27.0&  66.0 &  0.65      \\ \hline

     & \multicolumn{5}{c}{\textbf{DD dataset} }    \\ \hline
\multirow{3}{*}{ASEM vs. MoEL}    & Coh. &    \textbf{53.7}       & 33.3 & 13.0 & 0.39 \\
                                     & Emp. &   \textbf{37.3}        &  24.7 & 38.0 & 0.58         \\
                                     & Flu. &    \textbf{35.0}       & 25.3 & 39.7  &  0.42        \\ \hline
\multirow{3}{*}{ASEM vs.MultiTRS} & Coh. &    \textbf{46.7}       & 29 & 24.3 & 0.33        \\
                                     & Emp. &   \textbf{42.3}        & 18.7  & 39.0 & 0.46           \\
                                     & Flu. & \textbf{17.7} &16.0  & 66.3 &   0.52       \\ \hline

\multirow{3}{*}{ASEM vs. CASE}    & Coh. &    30.0     & \textbf{35.0} &  35.0& 0.37 \\
                                     & Emp. &   \textbf{29.0}      & 28.0 & 43.0 &  0.70      \\
                                     & Flu. &  2.0  & \textbf{14.0} &  84.0 &  0.88      \\ \hline
\end{tabular}
}
\caption{Human evaluation results($\%$) on the three aspects.}
\label{tab:HumanEvalResults}

\end{table}

\begin{table*}[]
\centering
\resizebox{\textwidth}{!}{%
\begin{tabular}{ll}
\hline
{\color[HTML]{000000} } \multirow{4}{*}{{\color[HTML]{000000} \textbf{Context}}} &
  {\color[HTML]{000000} S1: Two years ago i got diagnosed with ms.} \\ \cline{2-2} 
{\color[HTML]{000000} } &
  {\color[HTML]{000000} S2: What 'd ms ?} \\ \cline{2-2}  &

  {\color[HTML]{000000} \begin{tabular}[c]{@{}l@{}}S1: Its a disease that attacks the nervous system. It was really scary to find out i had it. I just \\ went blind in my right eye one day. Two days and several test later i found out i had ms.\end{tabular}} \\ \hline
{\color[HTML]{000000} \textbf{Speaker emotion:}} &
  {\color[HTML]{000000} Surprise} \\ \hline
{\color[HTML]{000000} \textbf{MoEL}} &
  {\color[HTML]{000000} I am so proud of you ! I hope you had a good experience.} \\
{\color[HTML]{000000} \textbf{MultiTrs}} &
  {\color[HTML]{000000} I hope it turns out well for you} \\
{\color[HTML]{000000} \textbf{ASEM}} &
  {\color[HTML]{000000} \textcolor{blue}{I 'm so sorry to hear that}. I hope it all works out for the best for you.} \\ \hline
{\color[HTML]{000000} \textbf{Golden}} &
  {\color[HTML]{000000} Well , I pray you get through it} \\ \hline
 &
   \\ \hline
    {\color[HTML]{000000} } \multirow{5}{*}{{\color[HTML]{000000} \textbf{Context}}} 
 &
 
  \begin{tabular}[c]{@{}l@{}}S1: When i lived in california my farmer friend and landlord had me get organized and have all \\ my valuables together in case of a big earthquake. He stockpiled such things as batteriesand water.\end{tabular} \\ \cline{2-2} 
 &

  \begin{tabular}[c]{@{}l@{}}S2: I can understand that as i live in Florida and we prepare for hurricanes each year with a plan \\ and emergency rations and supplies.\end{tabular} \\ \cline{2-2} &
  S1: It feels good to get everything together , doesnot it ? you feel as if you are ready for anything ! \\ \hline
\textbf{Speaker emotion:} &
  Anticipation \\ \hline
\textbf{MoEL} &
  Wow , that is a lot of deal . It is good that you have such a thing in your life \\
\textbf{MultiTrs} &
  That is good to hear. I am glad he 's able to help you out.  Are the best  in life \\
\textbf{ASEM} &
  That 's good that you \textcolor{blue}{were prepared}.  Can be so scary ! \\ \hline
\textbf{Golden} &
  \begin{tabular}[c]{@{}l@{}}Yes, you are correct. You can always hope for the best, But also plan for the worst. Being prepared \\ is a comfortable feeling\end{tabular} \\ \hline
\end{tabular}%
}
\caption{Two cases of ASEM and the baselines' responses. Blue words indicate ASEM's expected labels.}
\label{tab:Generated Responses}
\end{table*}




\begin{table}
\centering
\small
\begin{tabular}{ll}
\hline
\textbf{Context} &
  \begin{tabular}[c]{@{}c@{}}
  I never expected him to gift me that \\beautiful necklace for our anniversary!  \\
  \end{tabular}
\\
  \hline
\multicolumn{1}{l}{} & \multicolumn{1}{c}{\textbf{Responses}}                              \\ \hline
\multirow{1}{*}{\textbf{Sent.Only} }       & It's nice you have someone to look \\ &forward to it \\

\multirow{1}{*}{\textbf{Emo. Only}}   &   Well, I am glad you have a great time! \\&I am sure you can enjoy it. \\

\multirow{1}{*}{\textbf{Sent+Emo}}       &  
\textcolor{blue}{That is amazing}, you must have a very \\ &happy life!
\\

\hline
\end{tabular}%
\caption{Results for the Detection Effect of Sentiment and Emotion.}
\label{Sentiment and Emotion}
\end{table}

\subsection{Case Study}
Table \ref{tab:Generated Responses} displays the responses of ASEM and other baseline models in relation to two distinct emotional states. In the first scenario, MoEL failed to grasp the intended meaning of the user's statement, resulting in an irrelevant empathetic response with an incorrect emotion. Conversely, MultiTRS partially recognized that something was amiss while still generating a neutral and unhelpful response. ASEM effectively apprehended the gravity of the situation, acknowledging the user's emotions with an apology and offering emotional support with positive sentiment by responding with "I hope". In the second case, the user expressed anticipation towards a forthcoming event, and ASEM responded with "were prepared," validating that the model accurately captured the intended emotion and demonstrated a precise understanding of the meaning by replying, "Can be so scary". On the contrary, MoEL once again misunderstood the underlying significance of the last turn, and while MultiTRS acknowledged that the users received assistance from a farmer, it still deviated from the intended meaning, leading both baseline models to perceive it as something positive. Therefore, ASEM outperforms the baselines in terms of demonstrating empathy and understanding.

We have conducted additional experiments to illustrate the effects of using sentiment detection alone (Sent.), emotion detection alone (Emo.), and the combined use of emotion on top of sentiment detection (Sent+Emo). The results obtained from these experiments provide substantial support for the claims made in the paper as shown in Table \ref{Sentiment and Emotion}

\subsection{Ablation Analysis}
To verify the effectiveness of each model component, we conducted ablation studies, as shown in Table \ref{tab:Ablation study}:
\begin{itemize}
\item \textbf{W/O weighted concatenation(We. Conc.):} We observe a decline in context-appropriateness, lexical diversity, emotion classification, and Dist-2 score when disabling the weighted concatenation (We. Conc.) that gives more weight to the current turn. This indicates the importance of considering the current turn in generating appropriate and diverse responses.


\item \textbf{W One Enc-Dec:} Employing a mixture of experts with multiple encoders and decoders enhances the chatbot's ability to capture sentiment, generate emotionally aware responses, provide response variety, and improve context understanding. Ablating this component and employing only one encoder and one decoder leads to a significant drop in performance across these aspects. 
    
\item \textbf{W/O sentiment loss(S. Loss):} The absence of sentiment loss results in a decrease in macro F1 score, indicating the importance of incorporating sentiment loss in generating empathetic responses. The increase in perplexity suggests a decrease in coherence and difficulty in producing meaningful and relevant responses.

     
\item \textbf{W/O sentiment aware encoders(S. Enc):} Removing the sentiment-aware encoders impacts emotion classification, Dist-2 score, and PPL. Sentiment-aware encoders help the model developing effective feature representations and understanding diverse sentiment categories. The absence of this component hinders the model's ability to extend its understanding and generate appropriate responses.

\end{itemize}

Overall, the ablation studies confirm the effectiveness of each component in the ASEM model and highlight their contributions to improving various aspects of the model's performance.


\begin{table}[]
\centering
\resizebox{\columnwidth}{!}{%
\begin{tabular}{lllllll}
\hline
\textbf{Model} &
  \multicolumn{1}{c}{\textbf{PPL}} &
  \multicolumn{1}{c}{\textbf{BLEU}} &
  \multicolumn{1}{c}{\textbf{D-1}} &
  \multicolumn{1}{c}{\textbf{D-2}} &
  \multicolumn{1}{c}{\textbf{Cos.}} &
  \textbf{F1} \\ \hline
\textbf{ASEM} &
48.2 &
2.2 &
40.0 &
\textbf{6.7} &
\textbf{24.3} &
  \textbf{60.6} \\
\textbf{w/o We. Conc.} & 48.3 & \textbf{2.4} & 21.9 & 5.4 & 23.5 & 54.0\\
\textbf{One Enc-Dec}   &   \textbf{48.1}  &     2.1  &  20.0  &  5.0  &    23.7 & 53.0 \\
\textbf{w/o S. Loss}  & 57.2 & 2.1 & 40.0 & 6.5 & 23.0
& 53.3 \\
\textbf{w/o S. Enc.}  &   49.4 &   1.6 &  \textbf{42.4}    &  5.0 &    23.0                      
& 48.5  \\ \hline
\end{tabular}%
}
\caption{Ablation study of our proposed model ASEM using the ED dataset. The best results are highlighted in bold.}
\label{tab:Ablation study}
\end{table}

\section{Conclusion}
In this paper, we propose the ASEM model, which incorporates emotion analysis into sentiment analysis to improve the performance of an empathetic chatbot. We utilized the attention mechanism to focus on relevant aspects of the context while learning the feature representation. Experimental results on two empathetic datasets demonstrate that our model outperforms the baselines. Ablation analysis further confirms the importance of each component in our model. In future work, we plan to explore other attention mechanisms to investigate their impact on generating effective embeddings and improving chatbot performance.

\section*{Limitations}
Our work has some limitations. Firstly, the lack of a specific-task automatic metric makes it challenging to objectively measure and compare different chatbot models. Secondly, our model has a large number of parameters (around 74 million), which requires a significant amount of training data to achieve optimal performance. However, the available datasets used in our study may not be extensive enough, resulting in repetitive responses and limited diversity or creativity.




\section*{Ethics Statement}
The ED dataset \citet{rashkin-etal-2019-towards} is sourced from an open-source dataset and involves conversations focused on specified emotions conducted by crowd-sourced workers. The DD dataset \citet{li-etal-2017-dailydialog} was obtained from online platforms targeting English language learners. These conversations do not involve personal privacy concerns. We ensure anonymity in the human evaluation process to maintain ethical standards.

\section*{Acknowledgments}
This work was made possible by NPRP13S-0112-200037 grant from Qatar National Research Fund (a member of Qatar Foundation). The statements made herein are solely the responsibility of the authors.

\section{Bibliographical References}\label{sec:reference}

\bibliographystyle{lrec-coling2024-natbib}
\bibliography{lrec-coling2024-example}

\appendix

\section{Implementation Details}
\label{sec:appendixA}
For training our ASEM model, we utilized a virtual instance hosted on the Google Cloud Platform. This instance was equipped with 40 GB of RAM and a single NVIDIA A100 GPU unit. The number of trainable parameters in the model is around 57 million. Our model's training took approximately 20 hours to complete. Our approach involved leveraging preexisting code as a basis and building upon it to develop our solution using the PyTorch framework \citet{lin-etal-2019-moel}. To maximize training efficiency, we used a batch size of 16 and implemented the early stopping strategy. In addition, batch matrix multiplication is employed to improve the model's efficiency. To initialize the embedding, we used transfer learning techniques by employing 300-dimensional English-pretrained Glove embeddings \citet{pennington2014glove}. The model was optimized with the AdamW optimizer and trained for about 20K steps. Each encoder and decoder consists of 12 layers, with 10 attention heads per layer. In addition, we assigned a weight of 2.5 to the current turn to prioritize its importance. The weight matrices are initialized using Xavier initialization.

In the beam search decoding process employed during inference, the query passed to the decoder at each time step comprises the tokens that have been predicted in the preceding steps. Specifically, for the first step, the query is initiated with a start-of-sequence token, indicative of the beginning of a new response. For each subsequent step, the decoder receives the sequence of tokens that have been generated up to that point. This sequence is used to condition the prediction of the next token, enabling the model to generate a contextually relevant and coherent response progressively.

For emotion analysis, we produced a probability distribution across seven emotion categories. The choice of this specific number is a flexible hyperparameter within our methodology. This parameter allows us to customise the analysis based on the requirements of our study or the specific dataset we are examining. Consequently, while the ED dataset was mapped to 10 emotions and the DD dataset contains 6, a probability distribution over seven emotion classes is generated for the ED dataset and two experts for sentiments, while for the DD dataset, four experts represent different emotions with higher probabilities and three experts for sentiments. We ensured our model did not overfit by using regularization techniques and early stopping during training. These methods prevent the model from learning noise and patterns in the training data that do not generalize well.


\section{Dataset}
\label{sec:appendixDataset}
The conversations in the ED were conducted between pairs of human participants. One participant shared a personal narrative, and the other responded empathetically \citet{rashkin-etal-2019-towards}. The dataset provides annotations for 32 emotions, which we map to the eight basic emotions and two complex emotions defined by Plutchik's wheel of emotions, as shown in table \ref{tab:EDDataset}. The theory identifies various secondary emotions, including aggressiveness and optimism. Nonetheless, in our experimentation, we focused on just two secondary emotions (love and remorse) alongside the primary ones. This approach yielded optimal outcomes, as incorporating additional secondary emotions adversely impacted the accuracy of emotion detection. On the other hand, DailyDialog (DD) is a high-quality multi-turn open-domain English dialogue dataset \citet{li-etal-2017-dailydialog}. On average, there are around 8 speaker turns per dialogue, with around 15 tokens per turn. The DD dataset is already labelled with the six main emotions defined by Plutchik’s theory: anger, disgust, fear, happiness, sadness, surprise, and no emotion. Thus, we could not add more secondary emotions. We adhered to their predefined data splits to avoid overfitting. Specifically, the ED dataset is divided into 80\% training, 10\% validation, and 10\% testing splits, encompassing approximately 19,533, 2,770, and 2,547 dialogues, respectively. The DD dataset follows a similar split of 80\% for training and 10\% each for validation and testing, translating to roughly 11,118, 1,000, and 1,000 dialogues in each set.

\begin{table}[]
\centering
\resizebox{8cm}{!}{
\begin{tabular}{lll}
\hline
\textbf{Emotion} & \textbf{Sent} & \textbf{Original Emotions}              \\
\hline
Joy                               & 6568                             & \begin{tabular}[c]{@{}l@{}}excited/joyful/grateful/content/confident\end{tabular}         \\
Surprise                          & 3321                             & surprised/impressed                                                                         \\
Anticipation                      & 3734                             & \begin{tabular}[c]{@{}l@{}}anticipating/hopeful/prepared\end{tabular}                     \\
Love                              & 3412                             & \begin{tabular}[c]{@{}l@{}}sentimental/caring/nostalgic\end{tabular}                      \\
Trust                             & 3271                             & proud/trusting/faithful                                                                     \\
Anger                             & 5297                             & \begin{tabular}[c]{@{}l@{}}angry/annoyed/furious/jealous\end{tabular}                     \\
Disgust                           & 1270                             & disgusted                                                                                   \\
Fear                              & 5979                             & \begin{tabular}[c]{@{}l@{}}afraid/terrified/anxious/apprehensive \\ embarrassed\end{tabular} \\
Sadness                           & 5084                             & \begin{tabular}[c]{@{}l@{}}sad/lonely/devastated/disappointed\end{tabular}                \\
Remorse                           & 2314                             & guilty/ashamed                                                                              \\ \hline   
\textbf{10}      & \textbf{}       & \textbf{32}  \\                        \hline                                
\end{tabular} 
}
\caption{ED Dataset Statistics}
\label{tab:EDDataset}

\end{table}

\begin{table}

    \centering
    \begin{tabular}{>{\raggedright\arraybackslash}p{0.15\linewidth}>{\raggedright\arraybackslash}p{0.75\linewidth}} \hline 
        Label & Afraid\\ 
        Speaker & I've been hearing some strange noises around the house at night.\\ 
        Listener & oh no! That's scary! What do you think it is?\\ 
        Speaker & I don't know, that's what's making me anxious.\\ 
        Listener & I'm sorry to hear that. I wish I could help you figure it out\\ \hline
    \end{tabular}

    \caption{Example for one conversation from the ED dataset}
    \label{tab:ExampleED}

\end{table}

\begin{table}
    \centering
    \begin{tabular}{>{\raggedright\arraybackslash}p{0.75\linewidth}l}\hline
 Speaker/Listener&Emotion\\\hline 
        Speaker: "I was scared stiff of giving my first performance." & Fear\\ 
        Listener: " Were you? your performance was excellent." & Happiness\\ 
        Speaker:"Thank you for your kindly words." & Happiness\\ \hline 
 
    \end{tabular}
    \caption{Example for one conversation from the DD dataset}
    \label{tab:ExampleDD}
\end{table}

\section{More Analysis}
\label{sec:More Analysis}
\begin{figure*}
\centering
\subfloat{\includegraphics[width=0.7\linewidth,height=0.6\linewidth]{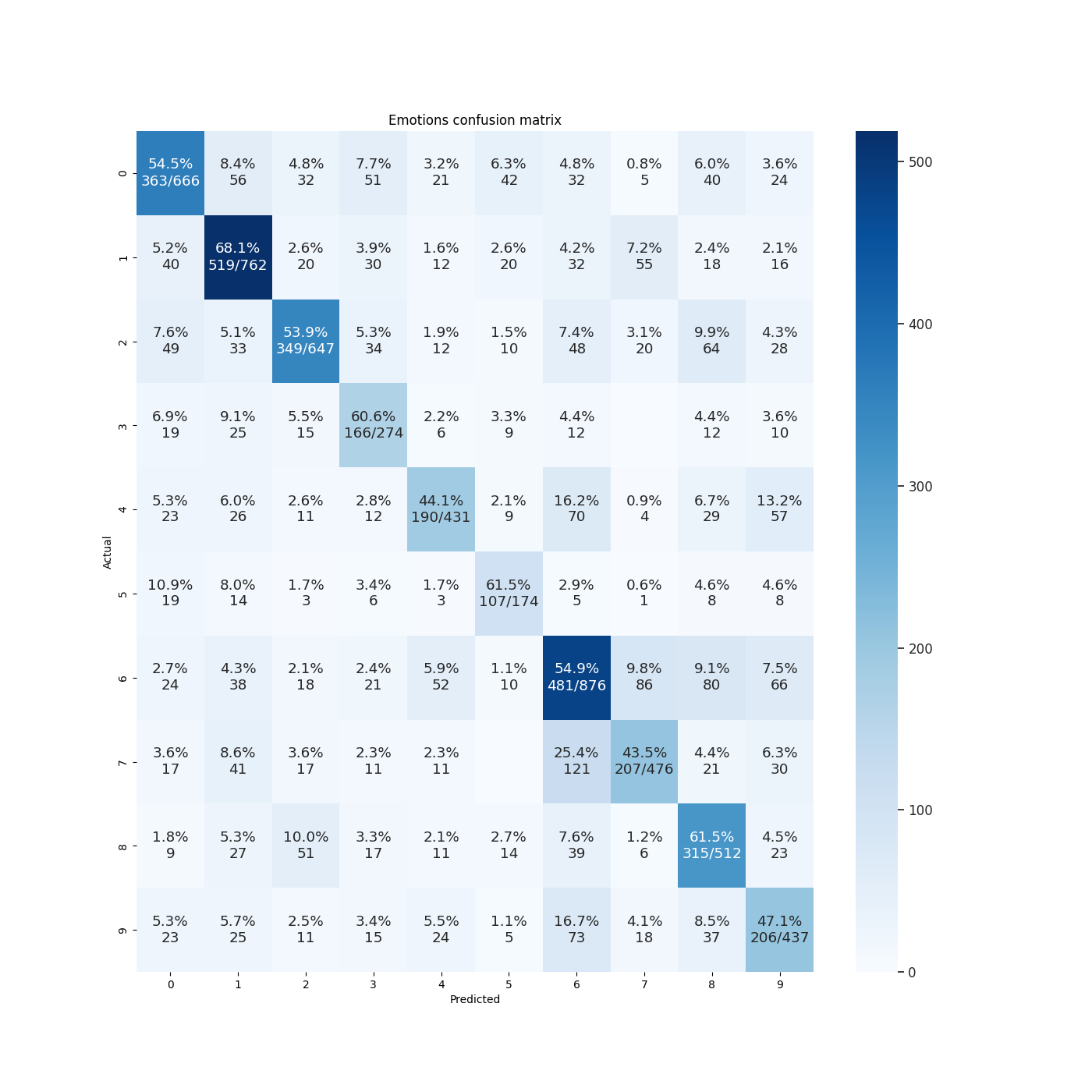}}
\hspace{-0.9cm}
\subfloat{\includegraphics[width=0.7\linewidth,height=0.6\linewidth]{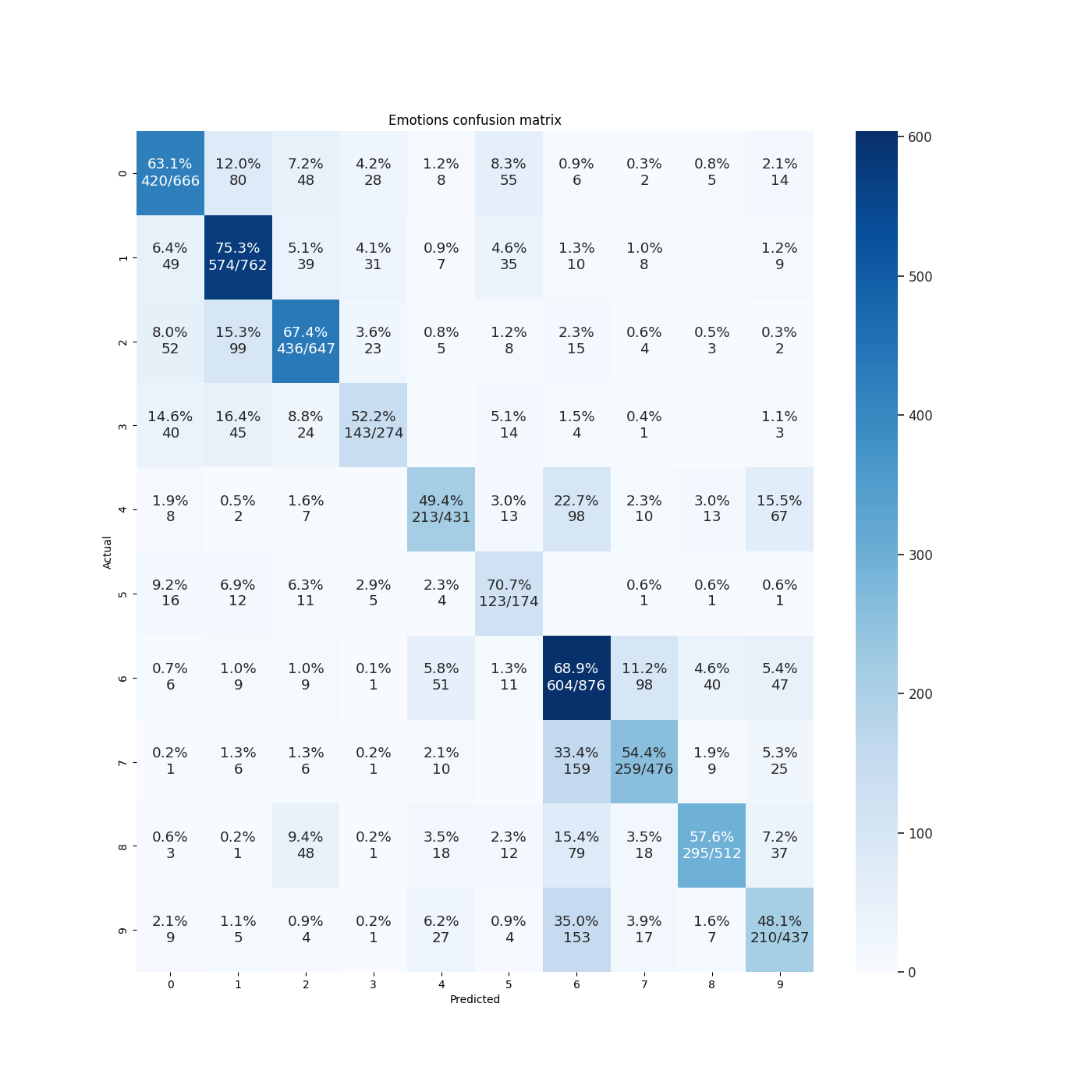}}
\caption{Confusion matrix for MoEL (top) and ASEM (down) models for emotion analysis using ED dataset (0: Anger, 1: Fear, 2: Sadness, 3: Remorse, 4: Surprise, 5: Disgust, 6: Joy, 7: Anticipation, 8: Love, 9: Trust ).}
\label{fig:Confusion matrix}
\end{figure*}

Figure \ref{fig:Confusion matrix} shows that ASEM improves performance in eight of ten emotion classes significantly. Additionally, Table \ref{Sentiment and Emotion comparsion} presents a comparative analysis of response generation based on sentiment, emotion, and their combination, along with explanations for the Integrated Approach (ASEM).

\begin{table*}
\centering

\small

\begin{tabular}{ll}
\hline
\textbf{Context} &
  \begin{tabular}[l]{@{}c@{}}
  Today, I think today is a day to learn, to hug, to smile, to make someone happy.  \\
  
  \end{tabular} \\

\textbf{Gold Response} &
  \begin{tabular}[l]{@{}c@{}}
  Yes, realistically, I wish it were possible to do that everyday, but general stresses of life get in the way.
    \end{tabular} \\
    \textbf{Emotion} & Anticipation 
\\
  \hline
\multicolumn{1}{l}{} & \multicolumn{1}{c}{\textbf{Responses}}                              \\ \hline
\multirow{1}{*}{\textbf{Sent.Only} }       & That is nice to hear. Do you have any plans for that? \\

\multirow{1}{*}{\textbf{Emo. Only}}   &   It is good to have something to look forward to. What will you do? \\

\multirow{1}{*}{\textbf{Sent+Emo}}       &  
You must be \textcolor{blue}{so excited!}, It is nice to have such a great time!
\\
\textbf{Explanation} & This response effectively combines positive sentiment with the anticipation emotion, \\&acknowledging the excitement and the intent to make the day special.

\\
\hline
\textbf{Context} &
  \begin{tabular}[l]{@{}c@{}}
  My daughter is doing such a great job juggling so many things after having a baby that was premature. \\
  
  \end{tabular} \\

\textbf{Gold Response} &
  \begin{tabular}[l]{@{}c@{}}

  that's good. I know that would be difficult.
    \end{tabular} \\
    \textbf{Emotion} & Trust 
\\
  \hline
\multicolumn{1}{l}{} & \multicolumn{1}{c}{\textbf{Responses}}                              \\ \hline
\multirow{1}{*}{\textbf{Sent.Only} }       & That is great. How old is she? \\

\multirow{1}{*}{\textbf{Emo. Only}}   &   That is great, how old is she ? \\

\multirow{1}{*}{\textbf{Sent+Emo}}       &  
That's so sweet. You must have a very happy life.
\\

\textbf{Explanation} & This response adds a layer of emotion by suggesting a sense of contentment and happiness \\&in life, which is more engaging but may not fully align with the context of trust and admiration \\&for the daughter's resilience.

\\
\hline

\textbf{Context} &

  I felt sooooo nervous going to my last job interview! Nerves must occur in proportion to how \\&badly you * want * something! \\

\textbf{Gold Response} &
  \begin{tabular}[l]{@{}c@{}}

  I know what you mean. Did you get the job?
    \end{tabular} \\
    \textbf{Emotion} & Fear 
\\
  \hline
\multicolumn{1}{l}{} & \multicolumn{1}{c}{\textbf{Responses}}                              \\ \hline
\multirow{1}{*}{\textbf{Sent.Only} }       & good for you. I am sure you will get one. \\

\multirow{1}{*}{\textbf{Emo. Only}}   &   Oh, i am sorry. I hope it goes well! \\

\multirow{1}{*}{\textbf{Sent+Emo}}       &  
\textcolor{blue}{Don't worry, you will be fine!}
\\
\textbf{Explanation} & 
This response directly addresses the user's fear, demonstrating emotion-supporting behavior \\&by providing reassurance.
\\
\hline

\textbf{Context} &
  \begin{tabular}[l]{@{}c@{}}
 I have been looking after my wife this week. She has the flu. \\
  
  \end{tabular} \\

\textbf{Gold Response} &
  \begin{tabular}[l]{@{}c@{}}

  Oh poor thing, is she at least getting better?
    \end{tabular} \\
    \textbf{Emotion} & Love 
\\
  \hline
\multicolumn{1}{l}{} & \multicolumn{1}{c}{\textbf{Responses}}                              \\ \hline
\multirow{1}{*}{\textbf{Sent.Only} }       & That is great! I hope she gets it. \\

\multirow{1}{*}{\textbf{Emo. Only}}   &   I am sorry to hear that. I hope she gets better. \\

\multirow{1}{*}{\textbf{Sent+Emo}}       &  
That's so nice to hear. I hope \textcolor{blue}{she feels better}.
\\
\textbf{Explanation} & 
This response, while not asking questions, expresses a positive sentiment and well wishes, \\&aligning somewhat with expressing care.
\\
\hline

\textbf{Context} &
  \begin{tabular}[l]{@{}c@{}}
 I was in total shock the night my fiancé proposed to me at a concert. \\
  
  \end{tabular} \\

\textbf{Gold Response} &
  \begin{tabular}[l]{@{}c@{}}

  That sounds like a great place to propose! How did it go?
    \end{tabular} \\
    \textbf{Emotion} & Surprise 
\\
  \hline
\multicolumn{1}{l}{} & \multicolumn{1}{c}{\textbf{Responses}}                              \\ \hline
\multirow{1}{*}{\textbf{Sent.Only} }       & Wow, how exciting! Where did you meet? \\

\multirow{1}{*}{\textbf{Emo. Only}}   &  That is awesome! Was it your birthday? \\

\multirow{1}{*}{\textbf{Sent+Emo}}       &  
\textcolor{blue}{Wow}, that is \textcolor{blue}{amazing news}. You must have been so \textcolor{blue}{excited!}
\\
\textbf{Explanation} & 
This response acknowledges the user's feelings of surprise and excitement.
\\
\hline

\textbf{Context} &
  \begin{tabular}[l]{@{}c@{}}
 Hello, I am very upset about the abuse of some people with their pets...Itis unacceptable \\
  
  \end{tabular} \\

\textbf{Gold Response} &
  \begin{tabular}[l]{@{}c@{}}

  I cannot imagine why people would do that. Animals are helpless.
    \end{tabular} \\
    \textbf{Emotion} & Anger 
\\
  \hline
\multicolumn{1}{l}{} & \multicolumn{1}{c}{\textbf{Responses}}                              \\ \hline
\multirow{1}{*}{\textbf{Sent.Only} }       & I am sorry to hear that. \\

\multirow{1}{*}{\textbf{Emo. Only}}   &  I am sorry to hear that. What happened? \\

\multirow{1}{*}{\textbf{Sent+Emo}}       &  
I hate when that happens. It's \textcolor{blue}{so annoying}.
\\
\textbf{Explanation} & 
This attempt to align with anger could be seen as demonstrating emotion-supporting \\&behavior, albeit less effectively.
\\
\hline

\textbf{Context} &
  \begin{tabular}[l]{@{}c@{}}
 I had to dogsit for my dad about a week ago. I had to clean up dog poop at least 10 times; it was awful. \\
  
  \end{tabular} \\

\textbf{Gold Response} &
  \begin{tabular}[l]{@{}c@{}}

  Ohh, cleaning dog poop is always been annoying, yet petting the dog is always amusing.
    \end{tabular} \\
    \textbf{Emotion} & Disgust 
\\
  \hline
\multicolumn{1}{l}{} & \multicolumn{1}{c}{\textbf{Responses}}                              \\ \hline
\multirow{1}{*}{\textbf{Sent.Only} }       & Oh no, that is awful. How did your dog get it back? \\

\multirow{1}{*}{\textbf{Emo. Only}}   &  Oh my gosh, that is so gross! I am so sorry to hear that! \\

\multirow{1}{*}{\textbf{Sent+Emo}}       &  
Wow, that is \textcolor{blue}{pretty gross}. What did you do?
\\
\textbf{Explanation} & 
This response acknowledges the user's feelings and asks a question that \\&encourages further sharing.
\\
\hline

\end{tabular}%
\caption{Comparative Analysis of Response Generation Based on Sentiment, Emotion, and Their Combination with Explanations for the Integrated Approach.}
\label{Sentiment and Emotion comparsion}
\end{table*}

\end{document}